# An interpretable clustering approach to safety climate analysis: examining driver group distinction in safety climate perceptions

Kailai Sun[1], Tianxiang Lan[1], Yang Miang Goh[1,], Sufiana Safiena[1], Yueng-Hsiang Huang[2], Bailey Lytle[3], Yimin He[3]

1 National University of Singapore, Singapore

2 Oregon Health & Science University, United States

3 University of Nebraska Omaha, United States

**Abstract**

The transportation industry, particularly the trucking sector, is prone to workplace accidents and fatalities. Accidents involving large trucks accounted for a considerable percentage of overall traffic fatalities. Recognizing the crucial role of safety climate in accident prevention, researchers have sought to understand its factors and measure its impact within organizations. While existing data-driven safety climate studies have made remarkable progress, clustering employees based on their safety climate perception is innovative and has not been extensively utilized in research. Identifying clusters of drivers based on their safety climate perception allows the organization to profile its workforce and devise more impactful interventions. The lack of utilizing the clustering approach could be due to difficulties interpreting or explaining the factors influencing employees' cluster membership. Moreover, existing safety-related studies did not compare multiple clustering algorithms, resulting in potential bias. To address these problems, this study introduces an interpretable clustering approach for safety climate analysis. This study compares five algorithms for clustering truck drivers based on their safety climate perceptions. It also proposes a novel method for quantitatively evaluating partial dependence plots (QPDP). Then, to better interpret the clustering results, this study introduces different interpretable machine learning measures (Shapley additive explanations, permutation feature importance, and QPDP). This study explains the clusters based on the importance of different safety climate factors. Drawing on data collected from more than 7,000 American truck drivers, this study significantly contributes to the scientific literature. It highlights the critical role of *supervisory care promotion* in distinguishing various driver groups. Moreover, it showcases the advantages of employing machine learning techniques, such as cluster analysis, to enrich the scientific knowledge in this field. The Python code used in this study is available at https://github.com/NUS-DBE/truck-driver-safety-climate.

# 1    Introduction

The transportation industry is a hotspot for workplace accidents and fatalities. In the United States, according to the Bureau of Labor Statistics (2020), transportation incidents are the most prevalent cause of work injuries, accounting for over 37.3% of workplace fatalities. Among transportation incidents, those involving large trucks are particularly dangerous. Accidents involving large trucks resulted in 5,005 deaths in 2019 (National Center for Statistics and Analysis, 2021). Trucking accidents are also especially severe. Even though only 3.3% of vehicles involved in crashes with injuries in 2019 were large trucks, they accounted for 9.8% of vehicles in fatal crashes (National Center for Statistics and Analysis, 2021). Thus, there is a need to continue improving trucking safety.

Safety climate has been identified by safety scientists as an important concept in accident prevention. Safety climate is defined as the shared perception among workers regarding their organization's policies, procedures, and practices with respect to the relative value and importance of safety (Griffin and Neal, 2000, Zohar, 1980, Zohar, 2000, Zohar, 2011). It may be understood as employees' perceptions of how important safe conduct is in their working behavior (Bhandari & Hallowell, 2022; Schüler & Vega Matuszczyk, 2022; Zohar, 1980). To reduce the risk and severity of workplace accidents, it is crucial to improve employees' shared perceptions of organizational safety policies and practices in order to enhance their safety performance and reduce safety hazards. A positive safety climate can prevent accidents (Kao et al., 2021) and improve safety performance (Huang et al., 2021). Safety climate has been shown to be a leading indicator of safety behavior and safety outcomes, such as in the trucking industry (Huang et al., 2013; Lee et al., 2019). Published meta-analytic studies suggest safety climate had a strong association with safety performance (Christian et al., 2009; Nahrgang et al., 2011). More specifically, utilizing data from the trucking industry, Huang et al. (2013) found that an increase in safety climate scores correlated with a reduction in lost workdays due to injury. Lee, Huang, Sinclair, et al. (2019) replicated these findings through a longitudinal design, highlighting the mediating role of safety behavior in the relationship between safety climate and injury severity. Furthermore, research has demonstrated that a positive safety climate in the trucking industry is linked to higher job satisfaction among truck drivers and reduced turnover rates (Huang et al., 2016). These findings underscore the critical importance of strengthening safety climate in the trucking sector.

Many studies on safety climate mainly focused on the identification of safety climate factor structure and the development of instruments to measure safety climate in companies (Huang et al., 2013; Kines et al., 2011; Oah et al., 2018). Though such findings may help companies and organizations to identify areas of improvement in their safety climate, interventions designed based on such findings are likely generic and unable to account for individual differences in safety climate perceptions. To tailor interventions to the characteristics of individuals, accurately clustering the respondents based on their safety climate perceptions is important.

Clustering is an unsupervised machine learning (ML) approach that aims to separate unlabeled datasets into groups of similar data points. In contrast, supervised machine learning methods rely on labeled datasets to predict classifications and values. Even though both supervised and unsupervised learning approaches are well-established, supervised learning is more commonly used in safety research in various industries (Bates et al., 2021; Goh et al., 2018; Goh & Chua, 2013; Goh & Sa'adon, 2015; Hegde & Rokseth, 2020; Jafari et al., 2019). One possible reason for the lack of interest in unsupervised learning algorithms is the difficulties in interpreting the clusters (Bertsimas et al., 2021). Many researchers interpret the clusters based only on the descriptive statistics of the clusters, leaving room for researchers' subjective judgments (Ma et al., 2021). In the context of safety climate research, low cluster interpretability impedes the identification of distinguishing factors of individual differences in the perception of safety climate. Better cluster interpretability helps to analyze what factors better distinguish drivers' safety climate perceptions. These reasons thus point to the need to use interpretable clustering in safety climate studies.

Another common drawback of recent unsupervised clustering studies in safety research is the lack of critical evaluation of results produced with different clustering algorithms. Despite the wide range of clustering algorithms available, existing studies tend to use only one or two algorithms (Jeong et al., 2022; Khanfar et al., 2022; Lombardi et al., 2023). This is not ideal since unsupervised clustering is a data-driven rather than theory-driven analysis, and different algorithms make different assumptions about the characteristics of the data. Using only one or two algorithms (e.g., DBSCAN, K-means) often converges to the local optimum (Fong et al., 2014; Yang et al., 2022), resulting in potential mistakes. To bridge the above knowledge gaps, this study addresses the following research issues:

1. How to categorize truck drivers based on their safety climate perceptions using multiple clustering algorithms.

2. How to derive the relative importance of safety climate factors using interpretable machine learning algorithms.

To the best of our knowledge, this is the first study to employ interpretable clustering approaches to analyze safety climate data, particularly within the trucking industry. The main contributions of this study are as follows:

1.  This study comprehensively compares five clustering algorithms (DBSCAN, K-means, Agglomerative, Mean shift, and BIRCH) to cluster the truck drivers based on their safety climate perceptions.

2. This study introduces different interpretable machine learning measures (Shapley additive explanations, permutation feature importance, partial dependence plots, and individual conditional expectation) to explain the clusters with the importance of different factors.

3. This study proposes a novel method for quantitative evaluation of partial dependence plots (QPDP). This method quantifies the importance of different factors in determining the cluster each sample belongs to, by using Kullback–Leibler divergence and mean square error. The overall code is available at https://github.com/NUS-DBE/truck-driver-safety-climate.

## 2 Literature Review

### 2.1 Safety Climate Studies in the Trucking Industry

Truck drivers are disproportionately affected by accidents resulting from various work conditions, such as fatigue, time pressure, and environmental conditions (Nahrgang et al., 2011). Long-distance truck drivers may experience isolation for extended periods of time, make real-time safety-related decisions, work without direct supervision, and react to emergencies without the assistance of colleagues (Huang et al., 2013; Lee et al., 2019). They may also face personal health-related risk factors, including mental stress, a lack of exercise, and sleep deprivation which may contribute to road accidents if not appropriately monitored (Crizzle et al., 2017). It is thus crucial to ensure that safety performance is maintained when drivers are unaccompanied. These conditions point to the need for a firm safety commitment at all levels of the company and a strong safety climate to minimize preventable accidents (Lee et al., 2019).

Safety climate has been found to be a robust predictor of safety performance and behavior (Christian et al., 2009; Huang et al., 2017; Nahrgang et al., 2011; Neal & Griffin, 2006). Huang et al. (2013) showed that safety climate perceptions correlate negatively with adverse safety outcomes and positively with safety behaviors. Delhomme and Gheorghiu (2021) found that factors related to trucking safety climate, such as supervisors' pressure for delivery deadlines and more safety training programs by the company, correlate with lower perceived stress, leading to fewer risky driving behaviors. Given that safety climate is closely related to safety performance, it is necessary for managers to establish and advocate for positive safety behaviors among employees (Nahrgang et al., 2011). Nevertheless, the above studies do not focus on differences among the drivers in the sample. Thus, their insights cannot be used to tailor the intervention measures to potential differences among subgroups of drivers.

Quantitative surveys are usually used to measure safety climate perceptions. Huang et al. (2013) developed a multi-level safety climate instrument for the trucking industry. Data from this instrument has since then been used to identify predictors of safety climate perception using a Bayesian network based on the leader-member exchange theory (Huang et al., 2021), and to identify factors indicative of high or low organizational commitment to safety climate (He et al., 2022), and so on. The instrument has also been applied outside the English-speaking world (Qu et al., 2022). Nevertheless, existing studies in safety climate have not adopted machine learning methods for analysis for identifying a comprehensive set of antecedents of safety climate and determining their relative importance (Bamel et al., 2020).

## 2.2    Machine Learning (ML) and Its Application in Safety Research

ML has been adopted in safety research in various industries such as healthcare (McCradden et al., 2020; Simsekler et al., 2021), building and construction (Poh et al., 2018), aviation (Puranik et al., 2020), and transportation (Mohammadnazar et al., 2021). ML algorithms may be broadly categorized into supervised and unsupervised learning. Supervised learning uses pre-labelled datasets to train models and then tests their performance in predicting the response variable, while unsupervised learning may categorize data into groups to facilitate the derivation of insights without distinguishing response variables from explanatory variables (Ayodele, 2010). Unsupervised clustering, in particular, may be used to categorize events and entities like incidents, individuals, companies or industries, providing insight for developing tailored interventions. Jeong et al. (2022) used DBSCAN to identify dominant traffic accident types and their associated factors. They found that road-user law violation is more closely related to pedestrian accidents and that environmental factors are not closely related to any cluster. Khanfar et al. (2022) used K-means to cluster driving behavior into conservative, standard and aggressive types based on vehicle kinematic parameters. They found that introducing a flashing green signal light makes drivers more conservative. Lombardi et al. (2023) used K-means to categorize occupational road accidents related to a landfill site based on road and weather conditions and proposed corresponding intervention measures. The common trait of the above studies is that they adopted only one clustering algorithm, which means potential biases of the algorithms cannot be controlled for. They also derived traits of the clusters with only descriptive statistics or simple statistical measures like ANOVA, without assessing how influential each variable is in distinguishing the clusters. This, thus, points to the potential of interpretable clustering methods in addressing the above gaps.

## 2.3    Interpretable machine learning

Interpretable machine learning has been developed in many fields. It is commonly used to conduct a post-hoc analysis of the importance of the features (predictors) on the target response in the supervised learning model (Watson, 2022). This allows the data user to understand the most important factor that distinguishes the categories of the outcome. For example, the Shapley additive explanations (SHAP) method constructs supervised learning models with all possible combinations of the features. It compares the predictions of these models with that of the original model with all the features (Lundberg & Lee, 2017). The attribution of each feature may then be calculated. Permutation feature importance (PFI) is another method to quantify feature importance. It is derived from the performance decrease when randomly permuting the values of a feature – the greater the performance decrease, the more important this predictor is to the model (Breiman, 2001). Partial dependence plot (PDP) shows the correlation between each feature and the target response averaged across all samples to reveal the marginal effect of the former on the latter (Friedman, 2001), which may be used to understand the feature importance. PDP visualizes how a specific input feature influences the target response while keeping other features constant (the marginal effect). The individual conditional expectation (ICE) plot also shows this correlation, only that it visualizes the correlation of the prediction with the feature for each sample per line rather than averaging the predictions across samples (Goldstein et al., 2015). Unlike quantitative measures like SHAP and PFI, PDP and ICE are usually evaluated qualitatively due to the lack of indices that can be derived from those plots. To overcome this, this study proposes the QPDP method to derive numerical feature importance values based on PDP and ICE.

The above methods are developed for supervised classification models. To interpret unsupervised clustering results, scientists from various fields have used the hybrid method of unsupervised clustering followed by supervised prediction (Bertsimas et al., 2021). This study refers to this method as the "cluster-then-predict" method. Specifically, supervised learning is used to predict the results generated by the unsupervised clustering, i.e., using the cluster as the label or target variable. Through this process, researchers can then calculate each feature's relative importance in the cluster membership prediction (Lau et al., 2022; Ma et al., 2021; Satre-Meloy et al., 2020). This explains why the samples end up in their particular clusters based on the values of their features. As indicated earlier, despite the increasing attention to interpretable ML, there are limited efforts in applying it to safety research. This study fills this gap in the literature.

In summary, this study aims to contribute to the scientific literature by identifying the role of safety climate perception in differentiating driver groups and exploring the benefits of interpretable clustering methods to advance our understanding of enhancing the safety of truck drivers.

## 3    Methodology

The overall methodology of this paper is divided into three phases as shown in Figure 1. In phase 1, we prepare the truck drivers' safety climate questionnaire responses for ML analysis. During this phase, the Huang et al. (2013) safety climate factors are calculated for each truck driver. Then, in phase 2, we use five algorithms (DBSCAN, K-means, agglomerative, Mean shift, and BIRCH) to cluster truck drivers based on their safety climate perception scores. In phase 3, the clusters that each driver belongs to are used as labels for training supervised ML models, with the safety climate perception scores as features, using the Gradient Boosting algorithm. This is the cluster-then-predict analysis explained earlier, and this analysis is repeated using the results of all five algorithms. A separate set of cluster results will be generated through an ensemble of the clustering results produced by all five algorithms. In the third step, for each set of cluster-then-predict analysis (corresponding to the five algorithms), its results will be analyzed with three interpretable ML methods: SHAP, PFI, and the proposed QPDP. Values of feature importance on target response will be derived from clusters produced by the five algorithms and based on the three interpretation methods (i.e., 15 values per feature). The 15 values will be assessed together to identify which features are more important than others. The flow of the methodology is illustrated in Figure 1.

*Figure 1 Overall methodology*

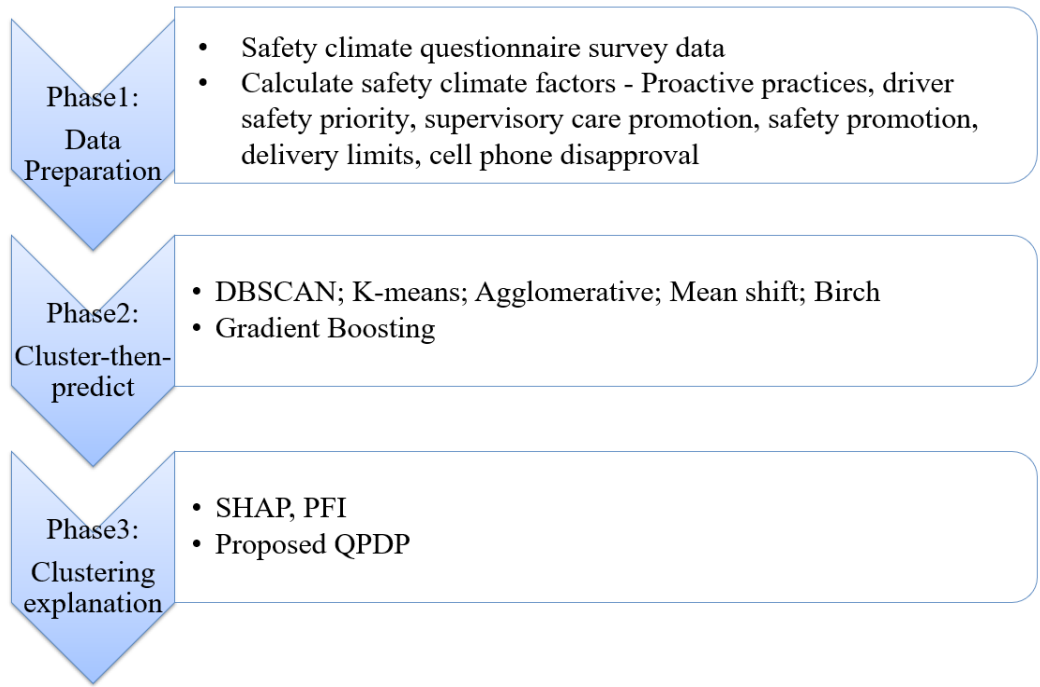

Phase1: Data Preparation
- Safety climate questionnaire survey data
- Calculate safety climate factors - Proactive practices, driver safety priority, supervisory care promotion, safety promotion, delivery limits, cell phone disapproval

Phase2: Cluster-then-predict
- DBSCAN; K-means; Agglomerative; Mean shift; Birch
- Gradient Boosting

Phase3: Clustering explanation
- SHAP, PFI
- Proposed QPDP

## 3.1 Data Preparation

This study uses the safety climate survey data collected by Huang et al. (2013). The survey was distributed to eight trucking companies via a web-based portal and conventional pen-and-paper method. The questionnaire consisted of organization- and group-level safety climate questions. Organization-level safety climate (OSC) relates to how the drivers perceive the company and top managers' safety promotion and management efforts. Group-level safety climate (GSC) relates to how drivers perceive their supervisors' or work groups' safety practices. A generic survey was first developed by Zohar and Luria (2005), then modified and validated for the trucking industry by Huang et al. (2013) and J. Lee et al. (2016). The survey consists of 94 items and uses a five-point Likert scale (1 = Strongly Disagree, and 5 = Strongly Agree) to measure workers' safety climate perception. Figure 2 shows the safety climate factors used in the survey.

Data was collected from 2011 to 2013. The dataset has a total of 7,474 entries. 398 incomplete data points with more than 10% missing values were removed to ensure the model generated is robust (Kang, 2013). We removed those data points rather than imputing the missing values for them because the dataset is large enough to allow the removal of those data points with marginal loss in power. The missing values of data points with less than 10% missing values were imputed using the median replacement function suggested by Lynch (2007).

In summary, the dataset contains 7,075 samples. Each sample has six features (OSC1-3, GSC1-3). All samples were used for the clustering, whereas for the supervised prediction, we used 70% of the data for training and 30% for testing. This is the first time that data has been used to study safety climate for truck drivers by employing various cluster analyses, comparing different clustering algorithms, introducing different interpretable machine learning measures to explain the clusters, and quantitatively evaluating partial dependence plots (QPDP).

*Figure 2 Factors in trucking safety climate scale developed by Huang et al. (2013)*

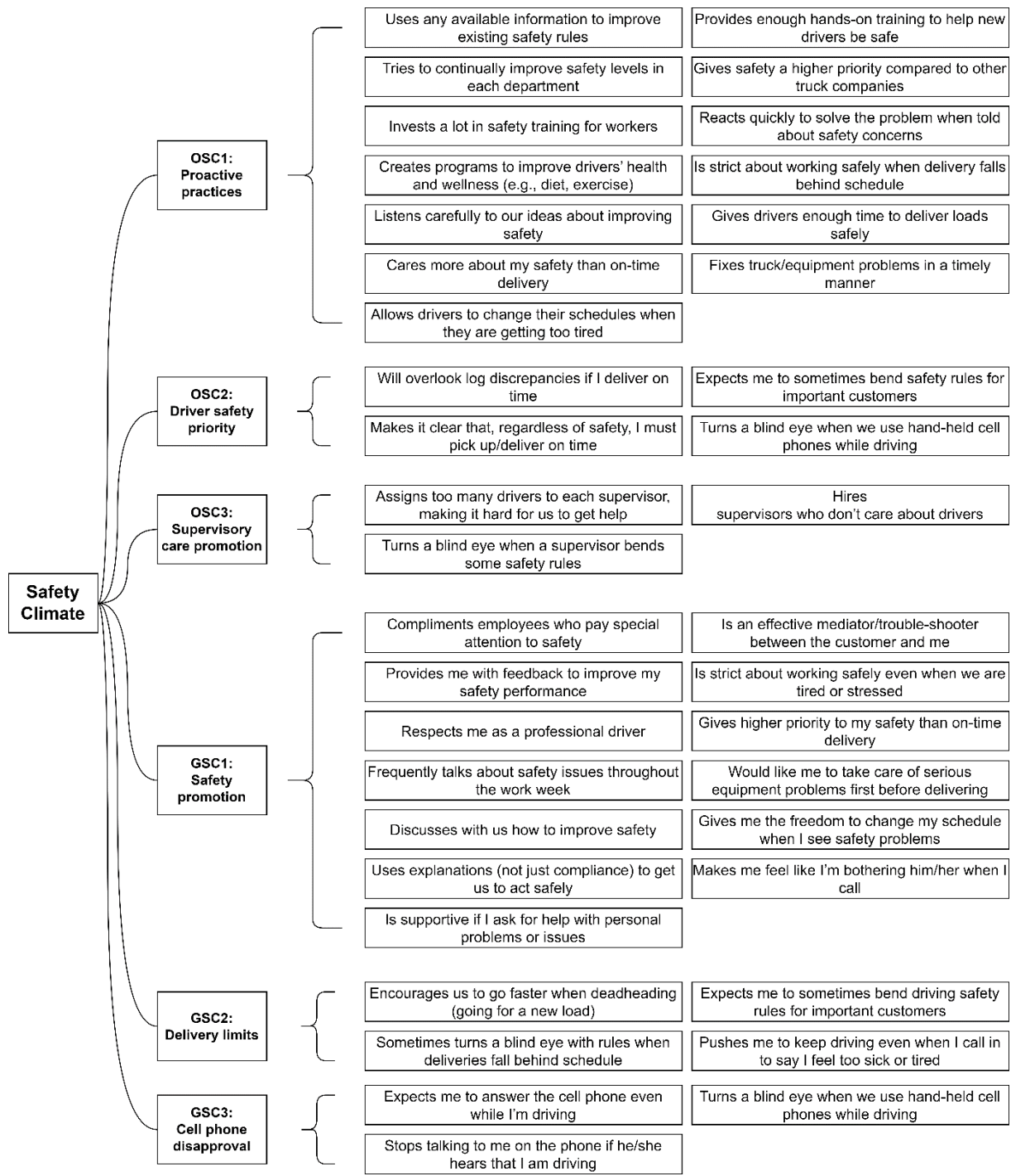

*Note:* OSC=organization safety climate, GSC=group safety climate. All responses for items with negative wordings were reversely coded.

## 3.2    Cluster-then-predict analysis

### 3.2.1    *Clustering*

As indicated earlier, five different algorithms were used to cluster the drivers according to their safety climate perception scores. The choice of these algorithms covers diverse types. They represent different categories of clustering algorithms, including centroid-based (K-means), density-based (DBSCAN, Mean shift), connectivity-based (Agglomerative) and compression-based (BIRCH) algorithms. Details of each algorithm are explained below.

K-means clustering separates the observations into k clusters such that the observations are distributed near the centroids of their clusters. The clustering is conducted such that the following error function is minimized:

$$E = \sum_{i=1}^{k} \sum_{x \in C_i} d(x, \mu(C_i)) \qquad\qquad ( 1 )$$

where C1, C2, ..., Ck are the k disjoint clusters, μ(Ci) is the centroid of each cluster and d(x, μ(Ci)) is the observation-to-centroid distance (Mohammadnazar et al., 2021). In this study, the k value is chosen by assessing which k value yields the optimal Silhouette Score and Calinski Harabasz Score. Silhouette Score measures the dissimilarity of each observation to other observations in its own cluster relative to its dissimilarity to observations in its nearest neighbor cluster (Kaufman & Rousseeuw, 1990). It ranges from 1 to -1, with 1 meaning the objects are well classified and -1 meaning the objects are misclassified (Everitt et al., 2011). Calinski Harabasz Score is the ratio of the sum of intra-cluster scatter (observation-to-centroid distance) and the value of inter-cluster separation (Duarte et al., 2010).

DBSCAN (Density-Based Spatial Clustering of Applications with Noise) clustering works based on identifying the "core points" and their neighbors. A data point is considered a core point if it has more than a minPts number of neighbors within a distance of ε from it. When different core points share neighbors, those core points and neighbors will be merged into a cluster. This process is iterated over every point. A data point not in any cluster is considered noise (Everitt et al., 2011; Schubert et al., 2017).

Agglomerative clustering is a type of hierarchical clustering. It works by fusing individual instances into groups successively. At each step, the individuals or groups closest to each other are fused. (Everitt et al., 2011). This study adopts Ward's method to determine how close the individuals or groups are to each other. It calculates the increase in the total within-cluster error sum of squares after the groups are fused, and the fused groups that generate the lowest increase in this error sum of squares are considered closest to each other (Everitt et al., 2011).

Mean Shift is a non-parametric, density-based iterative algorithm. It does not assume any pre-defined cluster shape, differentiating it from algorithms like k-means. It first selects random data points as starting cluster centers and identifies their neighbors within a pre-defined radius from the centers. Each cluster center is then iteratively shifted towards the densest regions in its neighborhood, until it does not move significantly (i.e., reaches convergence), and those are the positions of the final cluster centers (Comaniciu & Meer, 2002). This algorithm can adaptively determine the number of clusters rather than executing with a pre-specified number of clusters (Zhao et al., 2021).

BIRCH (Balanced Iterative Reducing and Clustering using Hierarchies) is a hierarchical clustering algorithm designed to handle large datasets efficiently. It reduces a large dataset into denser regions with statistical summaries called clustering features (CF). CFs are hierarchically organized into a CF tree, each leaf node containing multiple CFs. Leaf nodes are then grouped into final clusters with a traditional algorithm (K-means in this study) (Zhang et al., 1996). BIRCH is particularly useful for datasets with a large number of points, as it reduces the need for multiple passes over the entire dataset and optimizes memory usage.

### 3.2.2 *Supervised learning*

Supervised learning is adopted to construct predictive models to explain the clustering results. The target (predicted variable) is the cluster that the driver belongs to, while the predictors (features) are drivers' perception scores for each safety climate factor. This study uses Gradient Boosting Classifier, an ensemble machine learning algorithm. It builds a strong classifier by combining multiple weak classifiers (e.g., decision trees), through iterative training. The algorithm iteratively fits new weak classifiers to the errors made by previous iterations and combines their predictions with weighted voting to create the final prediction. This approach gradually improves model performance and accuracy by minimizing prediction errors.

## 3.3 Clustering explanation

### 3.3.1 *PDP*

The partial dependence plot (PDP) shows the marginal effect of one or more features on the target response of a supervised learning model (Friedman, 2001). A PDP can show the correlation between the target and a feature.

Let $X_I$ be the set of input features of interest and let $X_C$ be the complement of $X_I$. The partial dependence of the response $f$ at a point $x_i$ is defined as:

$$pd(x_i) = \int f(x_i, x_c) dx_{c,} \qquad\qquad (2)$$

where $f(x_i, x_c)$ is the response function for a given sample, whose values are defined by $x_i$ for the features in $X_I$ and by $x_c$ for the features in $X_C$. The partial function $pd(x_i)$ is the marginal effect of the given $x_i$ feature in the set $X_I$.

However, this integral is hard to compute. Instead, we estimate it by Monte Carlo methods (Altmann et al., 2020):

$$pd(x_i) \approx \frac{1}{N}\sum_{n=1}^{N} f\left(x_i, x_c^{(n)}\right), \qquad\qquad (3)$$

where N is the number of samples in our dataset. $x_c^{(n)}$ is the value of the n-th sample in the dataset for the features in $X_C$. The partial function $pd(x_i)$ is estimated by the marginal effect of the given $x_i$ feature in the dataset. For each $x_i$, this method requires a full pass over the dataset.

### 3.3.2 ICE

An individual conditional expectation (ICE) line is defined as a single $f\left(x_i, x_c^{(n)}\right)$ for the given feature $x_i$ (Molnar, 2022). An ICE line corresponds to a sample in the dataset. The average of the ICE lines corresponds to the partial dependence line.

### 3.3.3 Quantitative evaluation of PDP (QPDP)

Clustering truck drivers based on safety climate perception results in two clusters for all five algorithms used in this study. This will be explained in greater detail in section 4.1. For the purpose of interpretable clustering, this means the subsequent supervised learning models are all two-class classifications. Thus, we wish to develop a machine-learning model which can easily distinguish the samples from two classes. For example, in a linear classification model, the Sigmoid function can activate different values (0 or 1) based on the given features (0<, or >0). In this situation, the PDP is usually naturally distinguishable, i.e., a high-level value and a low-level value with different discriminated features.

To quantify and compare how the expected output depends on different features in the context of two-class classification, we propose QPDP, a PDP-based quantification method. It will be used together with SHAP and PFI to quantitatively evaluate each feature's importance in the classification model's target response. QPDP compares the normalized partial probability distribution functions, derived by converting the partial functions $pd(x_i)$ into probability distribution functions (PDFs), with the corresponding baseline PDFs. The baseline PDFs are step functions, representing the scenario when the value of a feature increasing beyond a certain threshold causes the target response to change from 0 to 1, or from 1 to 0. If a feature's normalized partial probability distribution function coincides with the baseline PDF, then it means the target response is maximally dependent on this feature. The distance between the normalized partial probability distribution and the baseline PDF, then, shows how much this feature falls short of having the maximal effect on the target response. This distance can thus be seen as a metric of the degree of dependency of the target response on this feature. The baseline PDF is adjusted to minimize its distance from the normalized partial probability distribution function:

$$\min_{\beta \in \{1,2,...M\}} D(nppd_\alpha || f_\beta) \qquad (\ 4\ )$$

where $\alpha$ represents the clustering method; $\beta$ represents baseline probability distribution functions (PDFs); $D$ represents the distance between the normalized partial probability distribution function ($nppd_\alpha$) and baseline PDF.

To achieve it, we define two baseline step PDFs:

$$f_a = \begin{cases} 0, & a > x \geq 0, \\ 1/(1-a), & 1 > x \geq a, \end{cases} \qquad (\ 5\ )$$

$$f_b = \begin{cases} 1/b, & b > x \geq 0, \\ 0, & 1 > x \geq b, \end{cases} \qquad (\ 6\ )$$

where $a, b \in (0,1)$ are the truncation points.

We adjust these truncation points to minimize the distance between the baseline step PDF and $nppd_\alpha$, and we use this distance to evaluate the dependence. To achieve it, we first normalize the $x_i$:

$$x_i = \frac{x_i - x_i^{min}}{x_i^{max} - x_i^{min}} \qquad (\ 7\ )$$

where the feature $x_i$ in the domain of definition is normalized to [0,1].

$$nppd(x_i) = \frac{pd(x_i)}{\int pd(x_i)dx_i} \qquad (8)$$

where $nppd(x_i)$ is the normalized partial probability distribution function for feature $x_i$. We introduce two metrics (Kullback-Leibler (KL) divergence, mean square error) to evaluate the distance between $nppd(x_i)$ and $f_a$ or $f_b$:

$$D_{KL}(f_a||nppd) = -\sum_{x_i \in X_I} f_a(x_i) \, log(\frac{f_a(x_i)+\varepsilon}{nppd(x_i)+\varepsilon}) \qquad (9)$$

$$MSE(f_a||nppd) = \sum_{x_i \in X_I}[f_a(x_i) - nppd(x_i)]^2 \qquad (10)$$

where $D_{KL} \geq 0$ and $MSE \geq 0$, measuring how one probability distribution is different from another. $\varepsilon$ is used to avoid 0 value. The above formulation also applies to $f_b$.

In summary, the proposed QPDP quantifies the dependency of target response on features based on PDP, through minimizing their distances and adjusting the truncation points $(a, b)$:

$$\min_{\substack{a \in (0,1) \\ b \in (0,1)}} \{D_{KL}(nppd_\alpha||f_a), D_{KL}(nppd_\alpha||f_b)\} \qquad (11)$$

$$\min_{\substack{a \in (0,1) \\ b \in (0,1)}} \{MSE(nppd_\alpha||f_a), MSE(nppd_\alpha||f_b)\} \qquad (12)$$

---

**Algorithm 1: Partial Dependence Quantification**

---

Input: samples $\{x^{(n)}\}_{n=1}^N$ for safety climate; feature dimension p

Output: best KL divergence array $D_{min}^*$

1. Initialize $T$xp array $D_{min}$ with $10^4$

2. **for** $1 \leq \alpha \leq T$ **do**

3.     **for** $1 \leq i \leq$ p **do**

4.         compute PD: $pd(x_i) \approx \frac{1}{N}\sum_{n=1}^N f\left(x_i, x_c^{(n)}\right),$ ;

5.         normalize PD: $nppd(x_i) = \frac{pd(x_i)}{\int pd(x_i)dx_i}$ ;

6.         search optimal $a^* \in \underset{a \in (0,1)}{\text{argmin}} -\sum_{x_i \in X_I} f_a(x_i) \, log(\frac{f_a(x_i)+\varepsilon}{nppd(x_i)+\varepsilon})$ ;

7.         search optimal $b^* \in \underset{b \in (0,1)}{\text{argmin}} -\sum_{x_i \in X_I} f_b(x_i) \, log(\frac{f_b(x_i)+\varepsilon}{nppd(x_i)+\varepsilon})$ ;

8.　　　　update $D_{min}^{\alpha,i} \in \min\{D_{min}^{\alpha,i}, D_{KL}(f_{a^*}||nppd(x_i)), D_{KL}(f_{b^*}||nppd(x_i));$

9.　　**end**

10.**end**

*Algorithm 1 Partial Dependence Quantification*

### 3.3.4　SHAP

The Shapley Additive exPlanations (SHAP) method was proposed to evaluate feature attributions (Lundberg & Lee, 2017). Shapley value is estimated by retraining the classification model on all feature subsets $S \subseteq X$, where $X$ is the set of all features. Each feature is assigned a value that indicates the impact on the model's prediction when the feature is included. This effect is calculated by training a model, denoted as $f_{S \cup \{i\}}$, with the $i$–th feature included, and another model, denoted as $f_S$, with the $i$–th feature excluded. Subsequently, the predictions from both models are compared, denoted as $f_{S \cup \{i\}}(x_{S \cup \{i\}}) - f_S(x_S)$ where $x_S$ represents the input feature values in the set S. This operation iterates through all possible combinations of features. The weighted average is obtained:

$$v_i = \sum_{S \subseteq X \setminus \{i\}} \frac{|S|!(|X|-|S|-1)!}{|X|!} [f_{S \cup \{i\}}(x_{S \cup \{i\}}) - f_S(x_S)] \quad (13)$$

### 3.3.5　PFI

Permutation Feature Importance (PFI) is a technique used in machine learning to gauge the importance of individual features (input variables) in our classification model. The method works by randomly permuting the values of each feature and observing the impact on the model's performance. A significant drop in performance upon permutation indicates that the feature in question is important for making accurate predictions. PFI is a model-agnostic method, meaning it can be applied to virtually any supervised learning algorithm to interpret the model's behavior. It provides an empirical way to measure feature importance and can be useful for both model interpretation and feature selection (Breiman, 2001).

# 4    Results

## 4.1    Unsupervised learning for cluster generation

### 4.1.1    K-means

To identify the optimal number of clusters, we used k values ranging from 2 to 11. The resultant Silhouette Score and Calinski Harabasz Score values were plotted against the k values, as shown in Figure 3. At k=2, the Silhouette Score and Calinski Harabasz Score are at their highest values. This suggests the optimal k value is 2.

*Figure 3 Silhouette Score and Calinski Harabasz Score plotted against k values for K-means clustering*

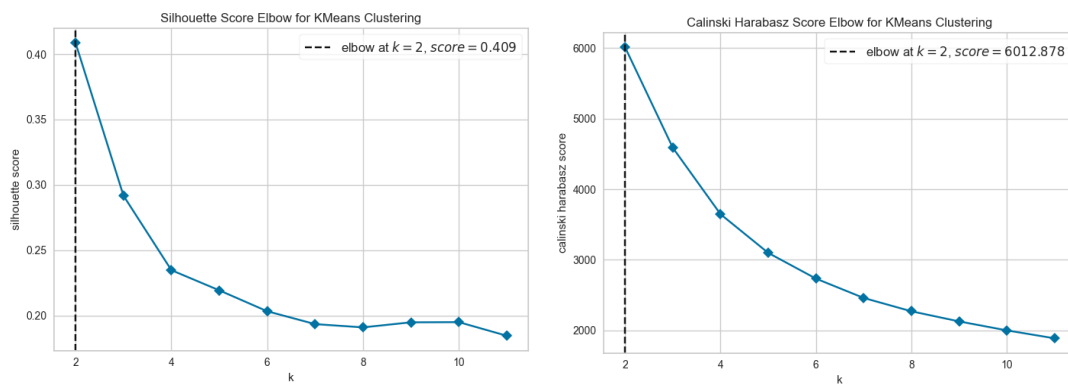

K-means clustering at k=2 produced two clusters of size 4,519 and 2,556, respectively. The means of drivers' safety climate perception scores for each factor in either cluster are shown in Table 1, and the number of drivers in either cluster is shown in Table 2.

*Table 1 Mean safety climate perception scores for each safety climate factor for clusters produced by five clustering algorithms*

| Algorithm | Cluster | Safety Climate factors | | | | | |
| | | OSC1 | OSC2 | OSC3 | GSC1 | GSC2 | GSC3 |
|---|---|---|---|---|---|---|---|
| **K-means** | Cluster 1 | 4.30 | 4.06 | 4.54 | 4.32 | 4.59 | 4.76 |
| | Cluster 2 | 3.23 | 2.86 | 3.37 | 3.03 | 3.21 | 3.43 |
| **DBSCAN** | Cluster 1 | 4.14 | 3.91 | 4.40 | 4.13 | 4.37 | 4.54 |
| | Cluster 2 | 3.34 | 2.88 | 3.37 | 3.13 | 3.38 | 3.60 |
| **Agglomerative** | Cluster 1 | 4.19 | 3.91 | 4.40 | 4.20 | 4.51 | 4.66 |
| | Cluster 2 | 3.14 | 2.81 | 3.31 | 2.89 | 2.90 | 3.19 |
| **Mean shift** | Cluster 1 | 4.35 | 4.14 | 4.60 | 4.37 | 4.62 | 4.78 |
| | Cluster 2 | 3.78 | 2.50 | 4.75 | 4.65 | 1.33 | 1.50 |
| **Birch** | Cluster 1 | 4.10 | 3.80 | 4.35 | 4.08 | 4.33 | 4.53 |
| | Cluster 2 | 2.89 | 2.67 | 2.83 | 2.66 | 2.83 | 2.91 |

Note: OSC1=Proactive promotion, OSC2=Driver safety priority, OSC3=Supervisory care promotion, GSC1=Safety promotion, GSC2=Delivery limits, GSC3=Cell phone disapproval.

*Table 2 Comparison of cluster sizes against different clustering methods*

| | K-means | DBSCAN | Agglomerative | Mean shift | Birch |
|---|---|---|---|---|---|
| **Cluster 1** | 4519 | 5121 | 5231 | 6109 | 5973 |
| **Cluster 2** | 2556 | 1954 | 1844 | 966 | 1102 |

### 4.1.2 DBSCAN

Following Rahmah & Sitanggang (2016), we determined the distance value $\varepsilon$ by calculating each instance's distance from its k-th nearest neighbor and plotting the distance from the lowest to the highest. The $\varepsilon$ value was then determined based on the k-distance at the point with the greatest slope change in the curve (i.e., the "knee point"). We plotted the graph using a k value of 12, based on twice the number of dimensions of the dataset (2*6=12). As shown in Figure 4, the optimal $\varepsilon$ value was 0.946, and we used this value for the DBSCAN analysis. Two clusters were produced, of size 5121 and 1954, respectively.

*Figure 4 Knee plot of distance to the 12th nearest neighbor for determining the ε value*

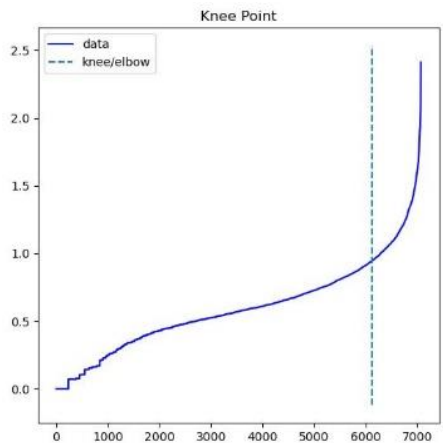

To verify that the drivers are indeed best split into two clusters, we iterated through the parameter of *min_sample* from 1 to 200, obtaining Silhouette and Calinski Harabasz Scores and the number of clusters. We then averaged the scores when the number of clusters produced was the same. In Figure 5, at k=2, the Silhouette Score and Calinski Harabasz Score were the highest. This means a two-cluster split is the best option. The average drivers' safety climate perception scores for each factor in either cluster are shown in Table 1.

*Figure 5 Silhouette Score and Calinski Harabasz Score plotted against k values for DBSCAN clustering.*

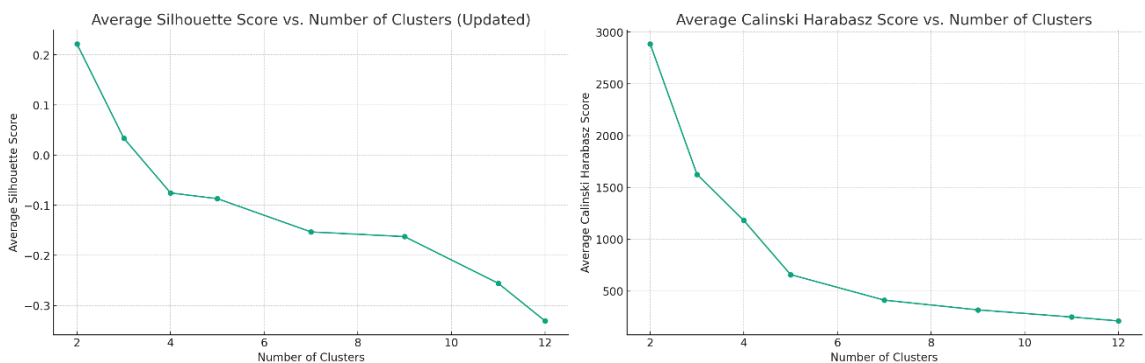

### 4.1.3 Agglomerative

To determine the optimal number of clusters, we used k values ranging from 2 to 11 and the resultant Silhouette Score and Calinski Harabasz Score values were plotted against the k values, as shown in Figure 6. At k=2, the Silhouette Score and the Calinski Harabasz Score are both at the highest value. This suggests the optimal k value is 2.

*Figure 6 Silhouette Score and Calinski Harabasz Score plotted against k values for Agglomerative clustering.*

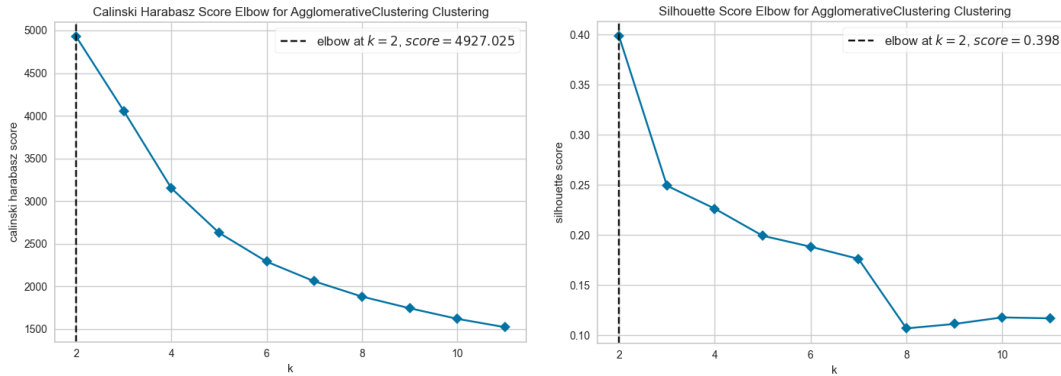

Agglomerative clustering with two clusters produced two clusters of size 5,231 and 1,844, respectively. The means of drivers' safety climate perception scores for each factor in either cluster are shown in Table 1.

### 4.1.4   Mean shift

Given that the mean shift algorithm does not require the number of clusters to be pre-defined, we iterated through the parameter of bandwidth from 0.1 to 0.3, obtaining Silhouette and Calinski Harabasz Scores and the number of clusters produced. We then averaged the scores when the number of clusters produced was the same. In Figure 7, at k=2, the Silhouette Score and Calinski Harabasz Score are at their highest values, which means a two-cluster split is the best option. Among the iterated bandwidth values that produce two clusters, we selected a value of 0.25 as the eventual bandwidth value to be used for clustering. The resultant two clusters were of size 6109 and 966, respectively. The means of drivers' safety climate perception scores for each factor in either cluster are shown in Table 1.

*Figure 7 Silhouette Score and Calinski Harabasz Score plotted against k values for mean shift clustering.*

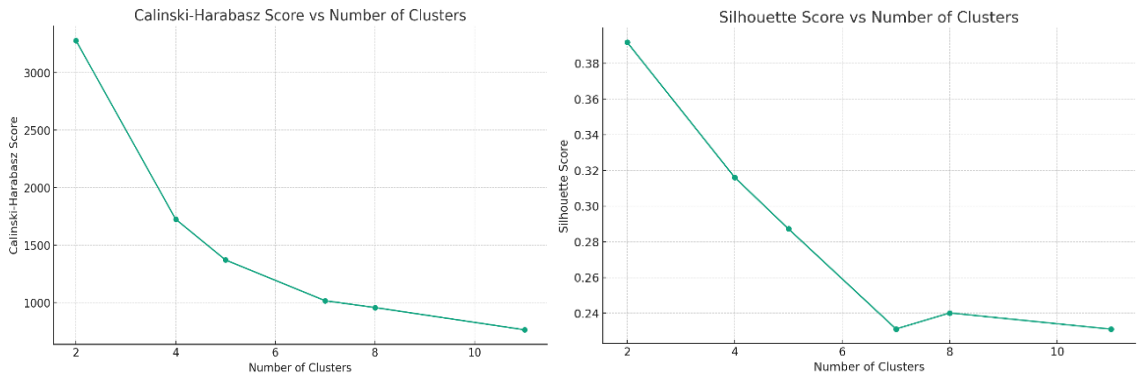

### 4.1.5   BIRCH

To determine the optimal number of clusters, we used k values ranging from 2 to 11; the resultant Silhouette Score and Calinski Harabasz Score values were plotted against the k values, as shown in Figure 8. At k=2, the Silhouette Score and the Calinski Harabasz Score are both at the highest value. This suggests the optimal k value is 2.

*Figure 8 Silhouette Score and Calinski Harabasz Score plotted against k values for BIRCH clustering.*

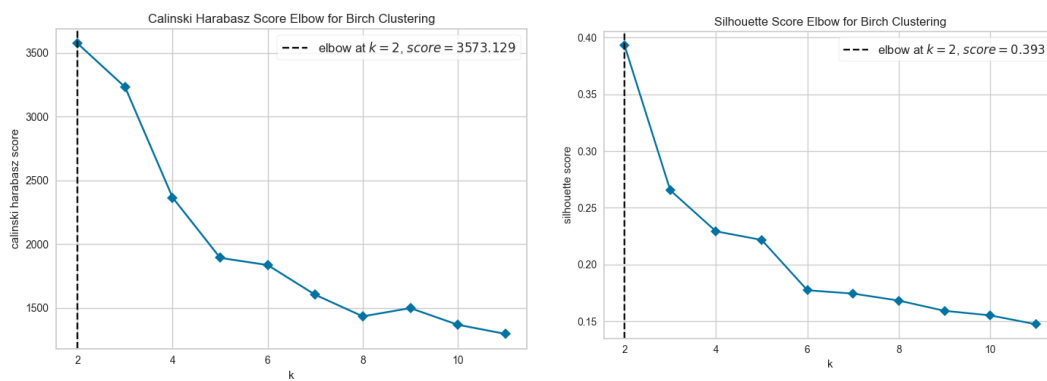

The BIRCH clustering produced two clusters of size 5,973 and 1,102, respectively. The means of the safety climate factor rating for either cluster are shown in Table 1.

In summary, through five different clustering algorithms, we found that five clustering algorithms agree that the optimal number of clusters is 2, through the analysis of Silhouette Score and Calinski Harabasz Score.

## 4.2 Cluster interpretation

For cluster interpretation, we developed the classification model using the Gradient Boosting algorithm to predict the cluster label from the features (safety climate perception scores for different safety climate factors). We trained one classification model for clustering results from each of the five algorithms. For each model, we split the dataset 7:3 for training and testing. Binomial and multinomial deviance were used as the loss function. For a fair comparison, we built the models for clusters from all five algorithms with the same set of hyperparameters: learning rate = 1.0; max depth of trees = 10; number of decision trees = 100. As shown in Table 3, all five models achieved an accuracy value ≥ 95%.

*Table 3 Accuracy of supervised learning models based on clusters generated by the five algorithms*

| DBSCAN | K-means | Agglomerative | Mean shift | BIRCH |
|--------|---------|---------------|------------|-------|
| 95%    | 97%     | 97%           | 98%        | 97%   |

### 4.2.1 QPDP

Based on the supervised learning models predicting the clustering results, the PDPs and ICE plots were created for the result of each clustering algorithm. Figure 9 shows the PDP and ICE for the prediction results based on clusters generated by DBSCAN. The rest of the PDPs and ICE plots are included in the appendix. The orange line in the figure shows the partial dependence (PD), while the blue lines show the ICE. We have 7,075 samples (i.e., 7075 blue lines in each image). PDP reveals the interaction between the target response and an input feature of interest, and one plot is produced for each feature (OSC1-3, GSC1-3).

We normalize the average line in Figure 9 to obtain the normalized partial probability distribution, shown as the blue line in Figure 10 and Figure 11. As explained earlier, the baseline step PDF is determined by minimizing its distance from the normalized partial probability distribution, based on KL divergence or MSE. Figure 10 and Figure 11 show the comparison between the normalized partial probability distributions and the baseline step PDFs determined based on KL divergence and MSE, respectively. The above figures were derived based on the results of DBSCAN clustering. Figures for other clustering algorithms are included in the appendix.

The KL divergence and MSE values are shown in Table 4 and Table 5. The lower value means a more significant dependence between the target response and the feature, and the lowest KL divergence or MSE for each model is bolded in the tables. It can be seen from both tables that the target response is more dependent on GSC3, OSC3, and OSC1. We find that the clustering results arising from DBSCAN and Mean shift are more dependent on the OSC1 feature. Results of K-means are more dependent on the GSC3 feature, and the result of Agglomerative method is more dependent on the GSC3 (Table 4) and OSC3 (Table 5) features. Lastly, the result of BIRCH is more dependent on the OSC3 feature. In summary, the GSC3 (cell phone disapproval), OSC3 (supervisory care promotion), and OSC1 (proactive practices) features tend to be more important in determining the clustering outcomes across the five clustering algorithms based on the evaluation of QPDP.

*Table 4 Comparison of KL divergence values for five clustering methods*

| Clustering methods | KL divergence | | | | | |
|---|---|---|---|---|---|---|
| | OSC1 | OSC2 | OSC3 | GSC1 | GSC2 | GSC3 |
| DBSCAN | **1.88** | 2.39 | 4.33 | 2.6 | 3.86 | 6.42 |
| K-means | 4.1 | 4.43 | 4.88 | 4.58 | 3.34 | **2.39** |
| Agglomerative | 4.3 | 6.14 | 2.57 | 4.65 | 2.56 | **2.44** |
| Mean shift | **0.04** | 6.39 | 15.91 | 16.91 | 8.27 | 4.06 |
| Birch | 5.93 | 5.38 | **1.35** | 5.73 | 2.2 | 2.16 |

*Table 5 Comparison of MSE values for five clustering methods*

| Clustering methods | MSE | | | | | |
|---|---|---|---|---|---|---|
| | OSC1 | OSC2 | OSC3 | GSC1 | GSC2 | GSC3 |
| DBSCAN | **0.021** | 0.034 | 0.077 | 0.026 | 0.054 | 0.103 |
| K-means | 0.087 | 0.14 | 0.14 | 0.121 | 0.063 | **0.061** |
| Agglomerative | 0.083 | 0.16 | **0.042** | 0.088 | 0.062 | 0.043 |
| Mean shift | **0.02** | 0.148 | 0.478 | 0.462 | 0.331 | 0.123 |
| Birch | 0.219 | 0.166 | **0.045** | 0.138 | 0.07 | 0.058 |

*Figure 9 ICE and PDP against different features (DBSCAN)*

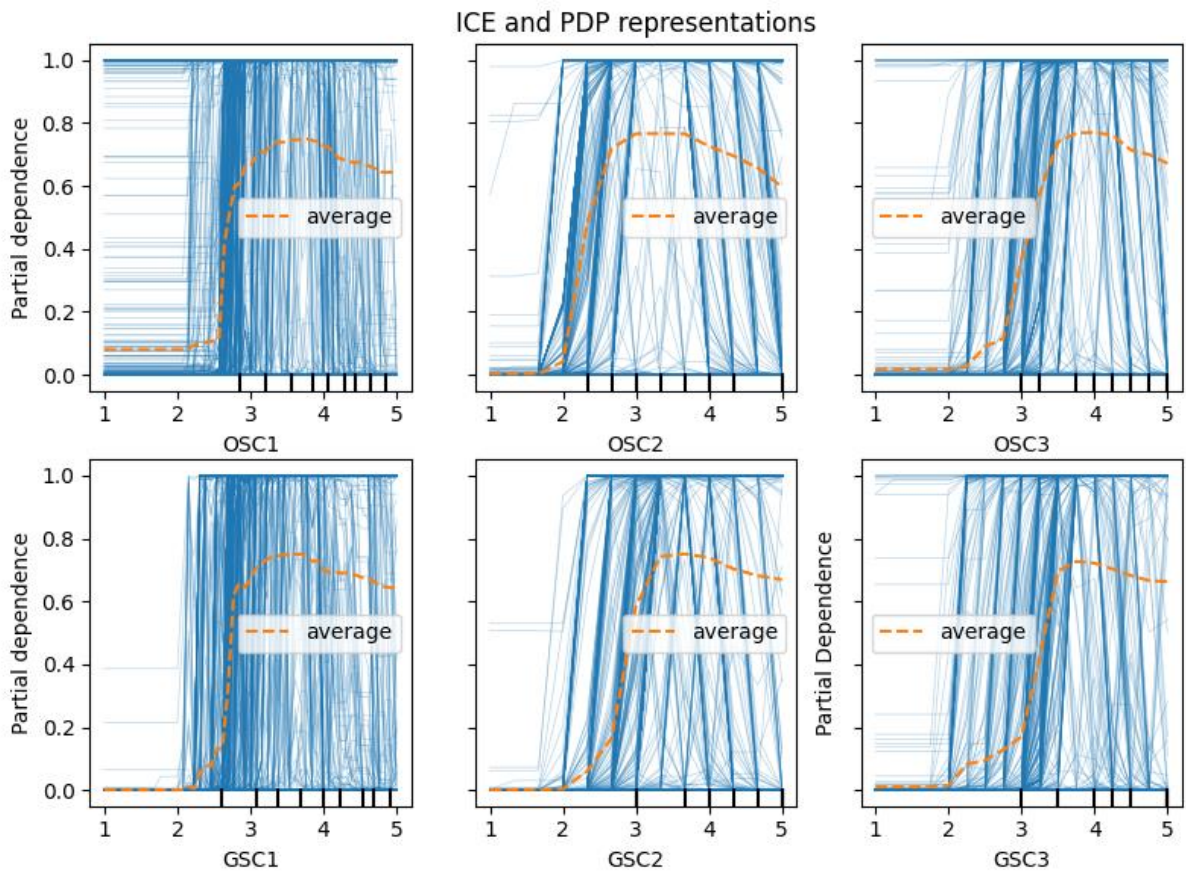

*Figure 10 Minimum KL divergence between normalized partial probability distribution and step function (DBSCAN)*

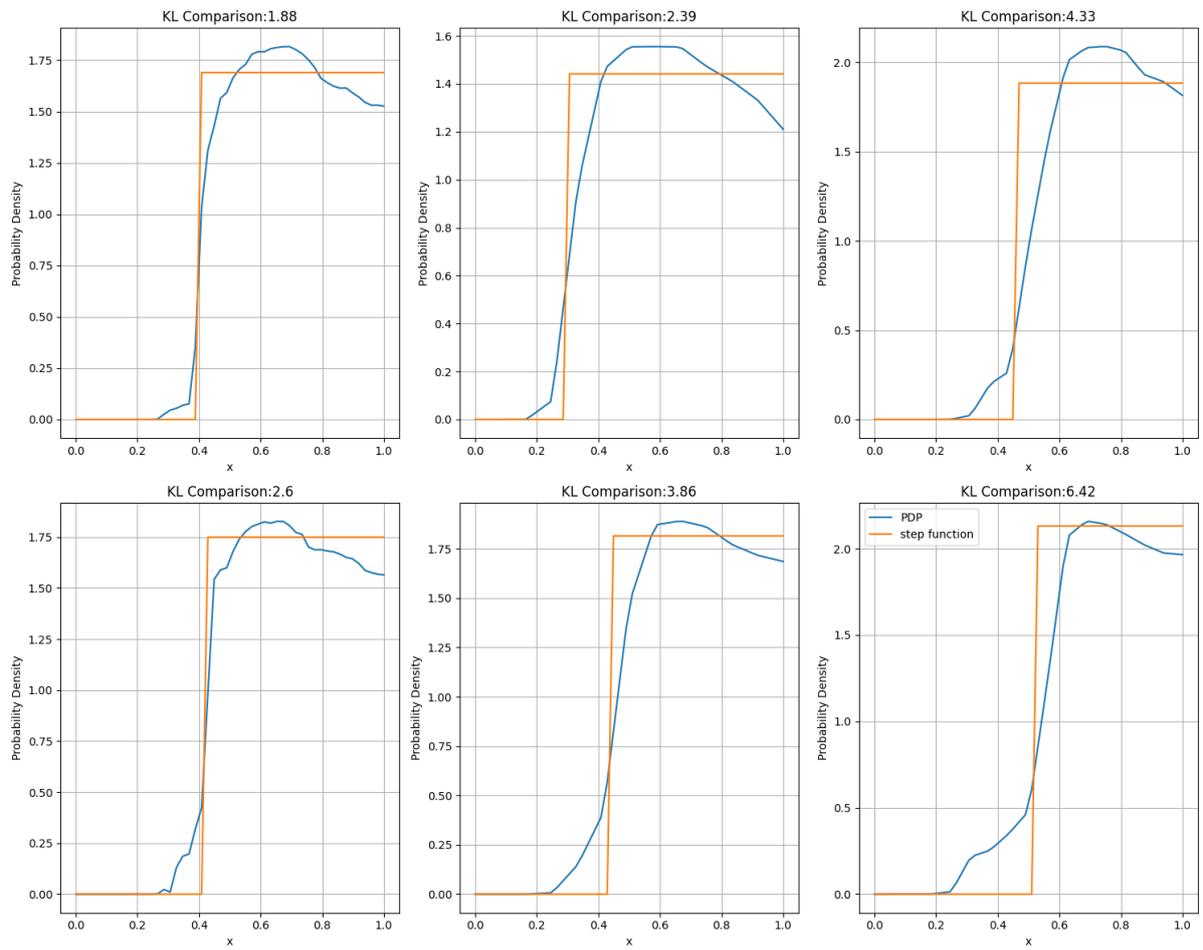

*Figure 11 Minimum MSE between normalized partial probability distribution and step function (DBSCAN)*

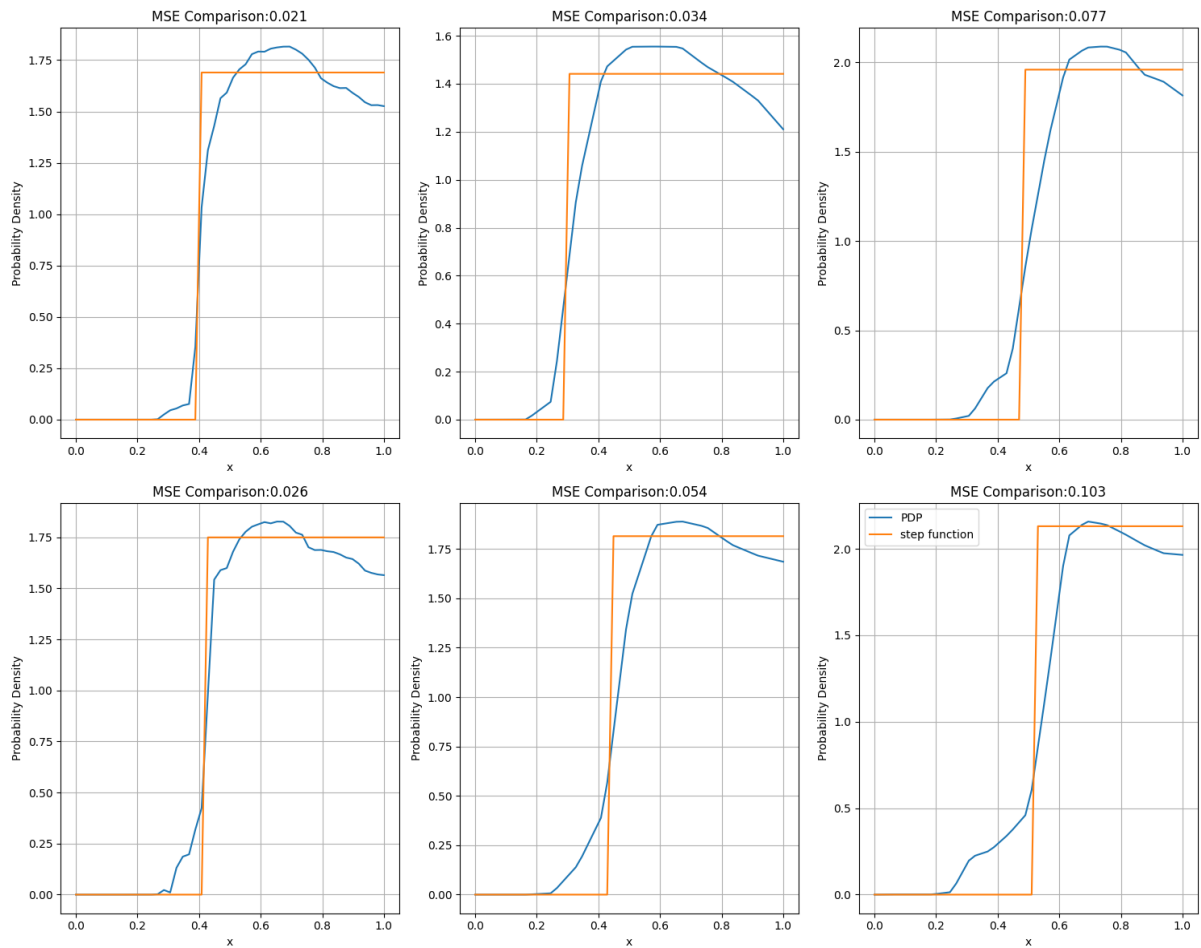

### 4.2.2   SHAP

In this section, we use SHAP methods to analyze the clustering results. The SHAP values using 1,000 samples on the testing dataset based on the DBSCAN clustering results are visualized in Figure 12. SHAP visualizations based on the results of other clustering algorithms are included in the appendix. The mean (|SHAP value|) based on the results of all five algorithms is shown in Table 6. It can be seen that the five clustering methods perform differently. In DBSCAN, OSC3 and OSC2 have bigger feature attributions to overall clustering; in K-means, GSC3 and GSC2 have bigger feature attributions; in Agglomerative and Mean shift, GSC3 and GSC2 have bigger feature attributions; in BIRCH, OSC3 and GSC3 have bigger feature attributions. DBSCAN's mean (|SHAP value|) based on clusters produced by DBSCAN has a small variance, indicating that DBSCAN can balance more features and information. In summary, GSC3 (cellphone disapproval), OSC3 (supervisory care promotion) and to a lesser extent GSC2 (delivery limits) have greater feature attributions than the rest. Thus, the SHAP method suggests that these three factors are more important to the clustering outcomes across the five clustering algorithms.

*Figure 12 SHAP value against different instances (DBSCAN)*

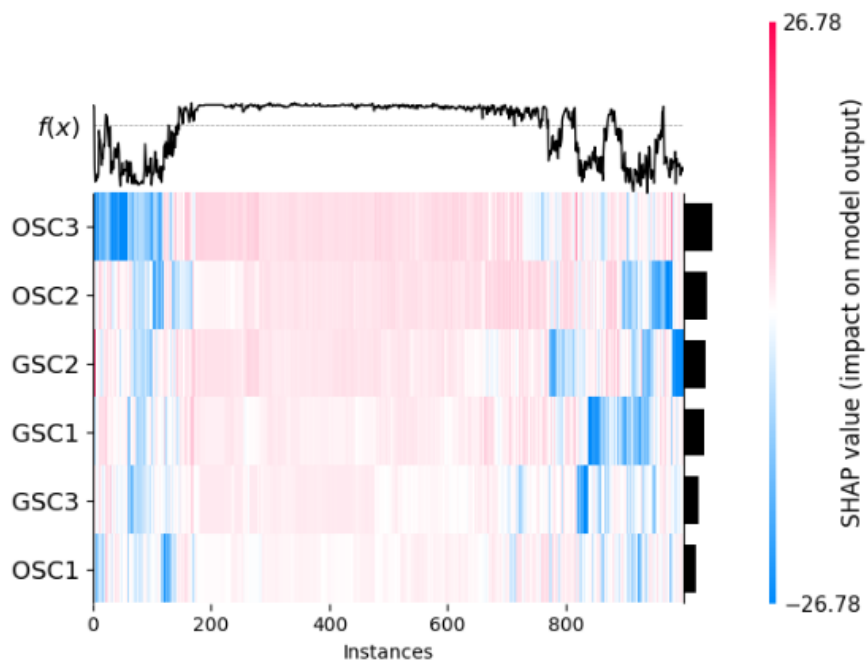

*Table 6 Comparison of Mean (|SHAP value|) against five clustering methods*

| Clustering methods | Mean (|SHAP value|) | | | | | |
|---|---|---|---|---|---|---|
| | OSC1 | OSC2 | OSC3 | GSC1 | GSC2 | GSC3 |
| DBSCAN | 2.24 | 4.06 | **4.98** | 3.36 | 3.7 | 2.51 |
| K-means | 3.17 | 3.77 | 3.53 | 3.46 | 5.19 | **6.2** |
| Agglomerative | 1.75 | 1.45 | 1.08 | 2.03 | **7.85** | 5.02 |
| Mean shift | 0.28 | 1.8 | 0.05 | 0.16 | 3.94 | **5.44** |
| Birch | 1.57 | 0.51 | **4.32** | 2.08 | 1.88 | 3.48 |

### 4.2.3 PFI

PFI scores for the supervised learning modules based on clustering results of five algorithms were generated over 30 times, the mean values of which are shown in Table 7. Overall, GSC2 is the most important distinguishing factor between clusters of drivers, followed by OSC3. In contrast, OSC1 and OSC2 consistently have lower importance on clustering outcomes than other features. In fact, GSC3 also has five high feature importance scores. In summary, OSC3 (supervisory care promotion) and GSC2 (delivery limits) have greater importance.

*Table 7 Comparison of PFI against five clustering methods*

| Clustering methods | PFI | | | | | |
|---|---|---|---|---|---|---|
| | OSC1 | OSC2 | OSC3 | GSC1 | GSC2 | GSC3 |
| DBSCAN | 0.417 | 0.508 | **0.742** | 0.614 | 0.644 | 0.700 |
| K-means | 0.130 | 0.161 | 0.185 | 0.178 | **0.305** | 0.211 |
| Agglomerative | 0.117 | 0.116 | 0.074 | 0.166 | **0.772** | 0.381 |
| Mean shift | 0.055 | 0.336 | 0.007 | 0.014 | **0.84** | 0.737 |
| Birch | 0.137 | 0.138 | **0.479** | 0.259 | 0.411 | 0.366 |

## 5    Discussion

This section addresses the two research issues indicated in the introduction.

## 5.1    How to categorize truck drivers based on their safety climate perceptions using multiple clustering algorithms.

Table 1 and Table 2 show that all five algorithms produced one larger and one smaller cluster, the size ratio between them ranging from 1.77:1 to over 6:1. The pattern of difference in each feature's means across the two clusters is largely consistent across the five algorithms, though the sizes and each feature's means fluctuate. For example, the mean value for GSC3 in cluster 2 ranges from 1.50 to 3.60, and that for GSC2 ranges from 1.33 to 3.38, as shown in Table 1. To balance out such differences produced by individual algorithms, we adopted a voting method to refine clusters for each driver, the results of which are shown in Table 8 and Table 9. For each driver, we computed its five cluster labels based on the five clustering algorithms, and then we determined the final cluster label through a majority vote. For example, if the labels are [1,1,2,1,1], 1 will be the final cluster label. This means that the drivers are clustered through an ensemble of multiple clustering algorithms. Compared with other clustering studies based on only one or two algorithms, the voting method based on five algorithms will be more accurate. This, in turn, means decision-makers can use the results to design personalized interventions more pertinent to workers' safety climate perception. For each driver, it will be more effective for supervisors to take customized action to care for and protect this driver. If an organization wishes to derive more detailed profiles of its employees, it may repeat the procedure in this study on the clusters to further categorize drivers within the selected cluster(s). This then allows the companies to further tailor intervention measures to smaller groups of employees to maximize effectiveness.

*Table 8 Mean safety climate perception scores for each safety climate factor based on voting results among the five algorithms.*

| Cluster | Safety Climate factors | | | | | |
|---------|------|------|------|------|------|------|
|         | OSC1 | OSC2 | OSC3 | GSC1 | GSC2 | GSC3 |
| Cluster 1 | 3.67 | 3.58 | 4.13 | 3.71 | 4.25 | 4.42 |
| Cluster 2 | 2.62 | 2.42 | 2.83 | 2.39 | 2.59 | 2.80 |

*Table 9 Number of drivers in each cluster based on voting results among the five algorithms.*

|  | K-means | DBSCAN | Agglomerative | Mean shift | BIRCH | **Voting** |
|--|---------|--------|---------------|------------|-------|--------|
| Cluster 1 | 4519 | 5121 | 5231 | 6109 | 5973 | 5473 |
| Cluster 2 | 2556 | 1954 | 1844 | 966 | 1102 | 1602 |

## 5.2   How to derive the relative importance of safety climate factors using interpretable machine learning algorithms.

### 5.2.1   Cluster-then-predict method

The cluster-then-predict method allows for the identification of features that influence the clustering outcome more strongly. The common practice of using only summary statistics or simple statistical methods like ANOVA is less accurate since, for any feature, the difference in its mean being statistically significant across clusters does not necessarily suggest that it is a more important factor for clustering. In contrast, the cluster-then-predict method allows the use of different methods to evaluate feature importance from the angle of dependence, attributions, and permutation importance, each capable of providing clear explanations of why the feature is more important. This, thus, makes the results more explainable and transparent, which has more practical value in empowering decision-making.

Comparing the three measures of feature importance on the prediction of clustering outcomes, QPDP suggests the most important features are GSC3, OSC3, and OSC1; SHAP suggests the most important ones are GSC3, OSC3; PFI suggests GSC2 and OSC3. All three measures agree on the importance of OSC3 (supervisor care promotion) is. This suggests that OSC3 is the most important distinguishing factor between the weak and the strong safety climate groups. The ability to identify this distinguishing factor points to the advantage of this study's methodology. As can be seen in Table 4 to Table 7, even though OSC3 can be identified as the distinguishing factor between the clusters, the relative level of importance of each factor still fluctuates widely across the five algorithms and three measures of feature importance. This, thus, points to the effect of the inherent biases and assumptions behind those algorithms and metrics. It allows us to identify the distinguishing safety climate factor with greater confidence by drawing strengths from multiple algorithms and metrics. This is important from a practical point of view. If a practitioner were to use only one clustering algorithm or metric of feature importance, one could have prioritized a different factor for improving safety climate. Consequently, the result may be more arbitrary or susceptible to the errors of the algorithm or metric chosen. We recommend future safety-related studies that use clustering to assist with prioritization to adopt this study's methodology, the overall flow of which is shown in Figure 1.

## 5.3    Interventions to improve safety climate

As explained earlier, OSC3 is the most important safety climate factor in the clustering results. Referring to Figure 2, the OSC3 safety climate factor is *Supervisory care promotion,* which includes these items: assigning too many drivers to each supervisor, making it hard for us to get help; turning a blind eye when a supervisor bends some safety rules; hires supervisors who don't care about drivers. These three items highlight the need to consider adjusting supervisor-to-driver ratios to ensure that supervisors can provide adequate support and assistance to drivers. It is imperative to cultivate a culture where safety is non-negotiable, and any deviations from safety protocols are promptly addressed, with clear consequences for disregarding safety established to promote adherence. Furthermore, there is a critical need to enhance the selection and training process for supervisors, actively seeking individuals who genuinely prioritize driver safety and well-being. Developing comprehensive training programs that prioritize safety over other operational and productivity concerns is essential. These items underscore the importance of safety taking precedence over competing operational demands and the need for supervisory oversight (Huang et al., 2013).

In short, this factor captures how companies manage the quantity and quality of their safety supervisors. This is consistent with studies suggesting the important role of the supervisor's responsibilities and competency in maintaining the company's good safety climate and workers' safe behaviors, especially when the supervisor is under administrative and productivity pressure (Peker et al., 2022; Tamene et al., 2022; Zahiri Harsini et al., 2020). In particular, Ahamed (2023) found that in the construction industry, top management ignoring or not monitoring supervisors' unsafe acts and bad safety practices is one of the most common errors. Those errors correspond well with the items in OSC3, suggesting that corporate mismanagement of supervisors poses obstacles to safety performance in the broader safety management context. Conversely, Tong et al. (2015) found that top management's empowering behavior to safety officers leads to better safety teamwork among the workers. This again points to the influence of decisions at the top management level on the performance at the supervisory level. Such insights allow the top management to identify priority areas of intervention to improve their companies' safety climate. Based on the findings above, they may, for instance, improve the quality and quantity of human resources to empower the supervisors, closely monitor the supervisors, and take seriously their bad practices to identify ways to improve without pinning the blame on them.

For companies to enhance their supervisors' quality and competency, proper training in the latter's leadership and mentorship skills is necessary, rather than simply promoting frontline workers to supervisors after a certain number of years of working. This is because the set of duties faced by supervisors and frontline staff are different (Greer, 2011). Existing studies have pointed to feasible ways to enhance the competency of supervisors. Bronkhorst et al. (2018) conducted a three-half-day safety leadership training for supervisors from healthcare organizations, covering both the theories of transformative leadership and roleplay exercises. Supervisors were then advised to apply the skills to their safety meetings. The intervention improved both the safety climate and safety behavior in those organizations. Zohar & Luria (2003) tested a scheme of providing supervisor feedback every week on how they interact with workers regarding safety-related matters. The results suggest that the intervention has also improved both safety climate and safety behavior. Such capacity building allows supervisors to understand their duties better and more confidently navigate the need for efficiency while maintaining safety as a top priority.

## 6 Limitation and future work

While the present study provides several important findings, we acknowledge the following limitations. First, the data in this study was collected from the United States only and focused on a sample of long-distance truck drivers. Discretion should be exercised when generalizing the findings to other regions where safety practices may differ. These findings would be most applicable in other regions where employees face similar unique working conditions to truck drivers. Second, the study focused on the cluster-then-predict approach of interpretive clustering due to its intuitiveness and that the supervised learning algorithms are well-established algorithms. Other forms of interpretive clustering can be further explored (Bertsimas et al., 2021). Future research can broaden data collection by including regions and countries with diverse safety practices, thus improving the generalizability of findings. Expanding the study's scope to encompass various industries with distinct working conditions will provide a more comprehensive understanding. Future research may also apply the interpretive machine learning methods to investigate safety climate at multiple levels, including team and organizational levels. Doing so could help researchers distinguish between different antecedents of safety climate across various levels and identify relevant intervention strategies tailored to specific levels.

## 7    Conclusions

Truck drivers continue to face unique and challenging work conditions; therefore, it is critical to have effective management that prioritizes safety. To reduce accidents, managers should design and implement appropriate interventions based on safety climate perception data to promote a safe working environment. Our study adopted an interpretable clustering approach. It minimizes the reliance on descriptive statistics alone and mitigates the potential for subjective judgments. We compared five clustering algorithms to cluster the truck drivers based on their safety climate perceptions. We proposed QPDP, a novel method for quantitative evaluation of partial dependence plots. We used it together with other interpretable machine learning measures (SHAP, PFI) to explain the clusters with the importance of factors on target response. The results indicated that the most important factor among clusters of drivers and their perception of safety climate was supervisory care promotion. The overall code is available at https://github.com/NUS-DBE/truck-driver-safety-climate.

## 8  Acknowledgement

The data in this study were collected while one of the authors worked at Liberty Mutual Research Institute for Safety. We thank the following team members for their invaluable assistance: Michelle Robertson, Susan Jeffries, Peg Rothwell, and Angela Garabet for data collection, analysis, and general assistance. In addition, we acknowledge the assistance of Michael Lim in conducting some of the pre-processing and analyses. This research did not receive any specific grant from funding agencies in the public, commercial, or not-for-profit sectors.